\documentclass[twocolumn,switch]{article}

\usepackage{preprint}

\usepackage[utf8]{inputenc}
\usepackage[T1]{fontenc}
\usepackage[numbers,square,sort&compress]{natbib}
\usepackage{amsmath,amssymb,amsfonts,amsthm}
\usepackage{graphicx}
\usepackage{xcolor}
\usepackage{booktabs}
\usepackage{multirow}
\usepackage{tabularx}
\usepackage{adjustbox}
\usepackage{algorithm}
\usepackage{algpseudocode}
\usepackage{float}
\usepackage{nicefrac}
\usepackage{microtype}
\usepackage{comment}

\usepackage{tabularx}
\usepackage{algorithm,amsmath}
\usepackage{algpseudocode}
\usepackage{multirow}
\usepackage{comment}
\usepackage{adjustbox}

\definecolor{linkblue}{rgb}{0.21,0.49,0.74}
\usepackage[
  colorlinks=true,
  linkcolor=linkblue,
  urlcolor=linkblue,
  citecolor=linkblue,
  anchorcolor=black
]{hyperref}
\usepackage[nameinlink,noabbrev]{cleveref}

\crefname{section}{Sec.}{Secs.}
\Crefname{section}{Section}{Sections}
\crefname{table}{Tab.}{Tabs.}
\Crefname{table}{Table}{Tables}
\crefname{figure}{Fig.}{Figs.}
\Crefname{figure}{Figure}{Figures}
\crefname{equation}{Eq.}{Eqs.}
\Crefname{equation}{Equation}{Equations}

\title{ViCrop-Det: Spatial Attention Entropy Guided Cropping for Training-Free Small-Object Detection}

\author{%
Hui Wang\textsuperscript{1} \quad
Hongze Li\textsuperscript{1} \quad
Wei Chen\textsuperscript{2} \quad
Xiaojin Zhang\textsuperscript{1,*}\\[0.5em]
\normalfont\small \textsuperscript{1}School of Computer Science and Technology, Huazhong University of Science and Technology\\
\normalfont\small \textsuperscript{2}School of Software Engineering, Huazhong University of Science and Technology\\
\normalfont\small Wuhan 430074, China\\[0.4em]
\normalfont\small
\texttt{chuxiatianyi@gmail.com} \quad
\texttt{lhz13012608937@outlook.com}\\
\normalfont\small
\texttt{lemuria\_chen@hust.edu.cn} \quad
\texttt{xiaojinzhang@hust.edu.cn}\\[0.4em]
\normalfont\small
\textsuperscript{*}Corresponding author \quad
}

\begin{document}

\lhead{\scshape Preprint -- ViCrop-Det}

\twocolumn[
  \begin{@twocolumnfalse}
    \maketitle
    \thispagestyle{empty}
    \begin{abstract}
Transformer-based architectures have established a dominant paradigm in global semantic perception; however, they remain fundamentally constrained by the profound spatial heterogeneity inherent in natural images. Specifically, the imposition of a uniform global receptive field across regions of varying information density inevitably leads to local feature degradation, particularly in dense conflict zones populated by microscopic targets. To address this mechanistic limitation, we propose ViCrop-Det, a training-free inference framework that introduces adaptive spatial trust region shrinkage. Inspired by the use of attention entropy in anomaly segmentation, ViCrop-Det leverages the detection decoder's cross-attention distribution as an endogenous probe. By utilizing Spatial Attention Entropy (SAE) to heuristically evaluate local spatial ambiguity, the framework executes dynamic spatial routing, allocating a fixed computational budget exclusively to regions exhibiting both high target saliency and high cognitive uncertainty. By shrinking the spatial trust region and injecting high-frequency localized observations, ViCrop-Det actively resolves spatial ambiguity and recovers fine-grained features without requiring architectural modifications. Extensive evaluations on VisDrone and DOTA-v1.5 demonstrate that ViCrop-Det yields competitive performance enhancements, consistently adding +1-3 mAP@50 to RT-DETR-R50 and Deformable DETR with a marginal 20-23\% latency overhead. On MS COCO, $AP_{S}$ improves while $AP_{M}/AP_{L}$ remains stable, indicating precise fine-scale refinement without compromising the global spatial prior. Under compute-matched settings, our adaptive routing strategy comprehensively surpasses uniform slicing baselines, achieving a highly optimized accuracy-speed trade-off.
\end{abstract}
    \vspace{0.35cm}
  \end{@twocolumnfalse}
]

\section{Introduction}
\label{sec:intro}
Despite their robust performance across standard benchmarks, global Transformer architectures continue to face a fundamental bottleneck in localizing microscopic targets within unconstrained, dense environments. This limitation is typically diagnosed empirically as "attention dilution," wherein the global receptive field disproportionately distributes softmax-normalized weights across expansive, low-information background patches~\cite{vaswani2017attention, carion2020end}. When a model encounters heavily downsampled, texture-degraded targets, the query embeddings fail to extract definitive localized spatial evidence. Consequently, the cross-attention probability mass diffuses, approaching a flat, uniform distribution across the spatial parameter space. This uniform spatial distribution inherently possesses a high Shannon entropy~\cite{shannon1948mathematical}, serving as a strong empirical indicator of the network's cognitive ambiguity and local spatial uncertainty. Forcing a definitive bounding box prediction from such a high-entropy state inevitably leads to persistent category confusion and extreme localization drift. While training-free uniform slicing methodologies (e.g., SAHI) attempt to recover these targets by artificially enlarging the pixel area~\cite{akyon2022slicing}, they operate on an assumption of spatial homogeneity. By blindly allocating computational resources to low-uncertainty, smooth background regions, these brute-force approaches incur severe computational overhead and fail to exploit the intrinsic spatial heterogeneity of natural images.

To address this mechanistic gap, we propose a training-free inference framework rooted in entropy-guided adaptive spatial routing. Inspired by recent findings demonstrating that spatial attention entropy can successfully segment unknown objects by measuring attention dispersion~\cite{lis2022attentropy}, we adapt and expand this principle to bounding-box detection by analyzing the endogenous cross-attention distribution of the Transformer decoder~\cite{vaswani2017attention}. Rather than relying on brute-force spatial slicing, our framework uses Spatial Attention Entropy (SAE) combined with attention intensity to create a joint ambiguity-saliency heuristic score. This allows the model to dynamically isolate localized conflict zones that suffer from peak spatial uncertainty but contain potential target evidence. Treating the inference process as a constrained computational resource allocation problem, our method executes \textit{entropy-guided adaptive spatial routing}: it actively redirects a strict computational budget exclusively to these heuristically identified ambiguous regions. By extracting localized, high-resolution observations strictly from these high-entropy patches and re-evaluating them, the network actively breaks the uniform spatial prior. This targeted injection of high-frequency visual evidence rectifies local feature representation and instantly resolves spatial ambiguity. Our primary contributions are summarized as follows:
\begin{itemize}
    \item We transition the challenge of small object detection from heuristic spatial scaling to entropy-guided uncertainty estimation. Building upon insights from anomaly segmentation~\cite{lis2022attentropy}, we introduce Spatial Attention Entropy (SAE) over aggregated decoder cross-attention as an effective empirical proxy for spatial cognitive ambiguity in Transformer detectors.
    \item We design a training-free adaptive spatial routing mechanism. By utilizing a joint ambiguity-saliency heuristic to dynamically allocate high-resolution computational budgets strictly to evidence-rich yet highly uncertain regions, the framework systematically bypasses the theoretical blindness and computational redundancy of uniform slicing paradigms.
    \item \textbf{Optimized Accuracy-Efficiency Trade-off:} Extensive evaluations on VisDrone~\cite{zhu2018visdrone} and DOTA-v1.5~\cite{xia2018dota} demonstrate that our uncertainty-driven approach yields consistent, scale-invariant performance gains over robust baseline architectures (RT-DETR~\cite{zhao2024detrs}, Deformable DETR~\cite{zhu2020deformable}). Furthermore, under compute-matched settings, it comprehensively surpasses uniform slicing baselines, validating the superiority of targeted, ambiguity-driven refinement.
\end{itemize}
\section{Related Work}

\textbf{Attention-based Inference Strategies in Vision Transformers.}
A range of vision transformer approaches leverage attention mechanisms for adaptive inference. AdaViT~\cite{meng2022adavit} dynamically adjusts the number of tokens during inference but requires retraining. DVT~\cite{wang2021not} prunes redundant tokens using attention scores, and DynamicViT~\cite{rao2021dynamicvit} adopts learnable token selection policies. Unlike ViCrop-Det's training-free approach, these methods require architectural modifications and retraining. PS-ViT~\cite{tang2022patch} uses patch slimming to enhance efficiency, not accuracy. TokenLearner~\cite{ryoo2021tokenlearner} learns spatial token aggregation, but is incompatible with fixed pre-trained models. Crucially, none of these methods leverage cross-attention from detection decoders to guide high-resolution region selection.

\textbf{Training-free Multi-scale Inference and Adaptive ROI Methods.}
Existing training-free strategies can be grouped into three main categories. (1) Uniform slicing: methods like SAHI~\cite{akyon2022slicing} and ASAHI~\cite{zhang2023adaptive} divide the image into fixed-size grids, treating all regions equally. This wastes computation on background regions while failing to focus on attention-dense areas. (2) Saliency-based selection: approaches such as CAM-based cropping~\cite{belharbi2021deep} and GradCAM-guided inference~\cite{zhang2019interpreting} rely on classification backbones, which do not align well with localization needs and often highlight object centers instead of boundaries. (3) Feature-level refinement: methods like FeatUp~\cite{fu2024featup} and LIIF~\cite{chen2021learning} upsample features post hoc but do not resolve attention dilution from the original forward pass. In contrast, ViCrop-Det uniquely utilizes decoder cross-attention entropy to identify ambiguous regions and selectively reprocess them at high resolution. Unlike ClusDet~\cite{yang2019clustered}, SNIP~\cite{singh2018analysis}, SNIPER~\cite{singh2018sniper}, LSKNet~\cite{li2025lsknet}, and BAFNet~\cite{song2024boundary}, which require training or focus on CNNs, our method is Transformer-specific and training-free.

\textbf{Attention Mechanisms for Object Detection.}
While attention visualization has been explored in tools like TokenCut~\cite{wang2023tokencut} and LOST~\cite{simeoni2021localizing}, few methods incorporate attention information directly into inference. QueryInst~\cite{fang2021instances} conditions queries on instance features, but requires training. Cascade R-CNN~\cite{cai2018cascade} refines detection proposals across stages but lacks attention awareness. Dynamic Head~\cite{dai2021dynamicb} adapts computation based on proposals yet still demands architectural modification. In contrast, ViCrop-Det extracts actionable signals from decoder attention: low entropy indicates confident localization needing no refinement, while high entropy reveals uncertainty---prompting selective high-resolution refinement without modifying the base model.

\noindent Our approach is orthogonal to token sparsification and dynamic-query
training~\cite{zheng2023less,huang2024dq}: they reallocate computation by
redesigning and training the detector, whereas ViCrop-Det is an inference-time,
model-agnostic router and can be combined with them in practice.
\section{Method}
\label{sec:method}

\begin{figure*}[t]
  \centering
  \includegraphics[width=\textwidth]{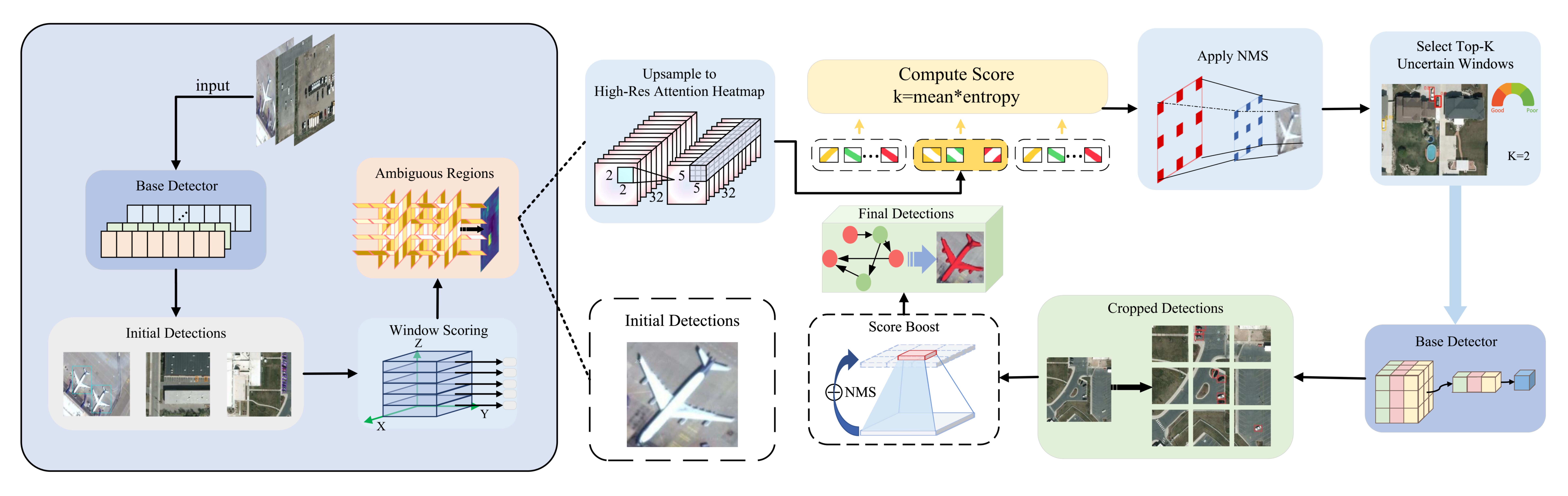}
  \caption{ViCrop-Det processing pipeline.}
  \label{fig:vicropdet}
\end{figure*}

The overall pipeline is illustrated in Fig.~\ref{fig:vicropdet}. In this section, we first formulate the spatial attention entropy as a metric to quantify spatial ambiguity (Sec. 3.1). Subsequently, we introduce the joint ambiguity-saliency heuristic scoring mechanism (Sec. 3.2). Finally, we detail the entropy-guided adaptive spatial routing and the corresponding variance reduction process (Sec. 3.3 to 3.7).

\subsection{Spatial Attention Entropy as an Ambiguity Indicator}
Standard Vision Transformers impose a uniform global receptive field across the spatial parameter space. While effective for macroscopic structures or homogeneous backgrounds, this global prior becomes critically sub-optimal in dense, heterogeneous conflict zones containing microscopic targets. To evaluate the resulting spatial ambiguity without introducing external heuristic modules, we analyze the decoder cross-attention distribution.

Let the aggregated, normalized cross-attention map over the spatial grid $\mathcal{P}$ be denoted as a probability distribution $p_i$, where $i \in \mathcal{P}$. This distribution essentially represents the network's posterior spatial hypothesis regarding the target's location. In a theoretically optimal, high-confidence regime, $p_i$ approaches a Dirac delta distribution centered precisely on the target, maximizing the semantic signal. However, in regions afflicted by heavy feature downsampling and severe object entanglement, the probability mass disperses. We quantify this spatial dispersion by computing the Spatial Attention Entropy (SAE). Specifically, we first aggregate the attention across layers, heads, and queries to obtain a single attention score per key:
\begin{equation}
\label{eq:aggregated_attention}
\mathbf{A}[i] = \frac{1}{L\,N_h\,N_q}\sum_{l=1}^{L}\sum_{h=1}^{N_h}\sum_{q=1}^{N_q} \bigl(\mathbf{A}_{l,h}\bigr)_{q,i},\quad i\in\mathcal{P}.
\end{equation}
We then normalize it to a probability distribution:
\begin{equation}
\label{eq:prob_norm}
p_i=\frac{\mathbf{A}[i]+\varepsilon}{\sum_{j\in\mathcal{P}}\bigl(\mathbf{A}[j]+\varepsilon\bigr)}, \qquad \varepsilon=10^{-8},
\end{equation}
and calculate the normalized Shannon entropy:
\begin{equation}
\label{eq:sae}
\hat{H}_s = - \frac{1}{\log N_k} \sum_{i \in \mathcal{P}} p_i \log p_i
\end{equation}
where $N_k$ is the total number of spatial keys. In this formulation, $\hat{H}_s \in [0, 1]$ serves as an effective, endogenous indicator of the model's spatial \textit{cognitive ambiguity}. A highly dispersed attention map yields a high $\hat{H}_s$, indicating that the local feature representation is severely entangled or lacks discriminative power, rendering the model incapable of definitively resolving target boundaries amid dense background interference. We reshape $\mathbf{A}$ to the key grid $(H_k,W_k)$ and bilinearly upsample to $(H,W)$, yielding a high-resolution attention map $\mathbf{H}\!\in\![0,1]^{H\times W}$, followed by min--max normalization.

\paragraph{Thresholds.}
We distinguish two entropy thresholds: a \emph{global} threshold $\tau_g$ applied to the image-level entropy $\widehat H_s$ (for early-exit gating), and a \emph{window-level} threshold $\tau_w$ applied to the per-window entropy $\widehat H_s(w)$ (for candidate selection). We also denote by $\mu$ the attention-intensity threshold applied to $m(w)$. Unless otherwise stated, we use $\tau_g=\tau_w=0.7$ and $\mu=0.3$ by default, selected by a small grid search on the validation splits of DOTA-v1.5 and VisDrone, and then frozen across all detectors/datasets to balance accuracy and overhead.

\paragraph{Early-exit gating.}
To avoid unnecessary refinement on easy images, we apply a global gate:
\begin{equation}
\label{eq:gate}
\text{proceed} = \begin{cases}
1, & \bigl(\widehat H_s \ge \tau_g\bigr) \ \land\ \bigl(\operatorname{mean}(\mathbf{H}) \ge \mu\bigr),\\
0, & \text{otherwise}.
\end{cases}
\end{equation}
Intuitively, low entropy indicates sharply focused attention (little ambiguity), while very low mean attention suggests insufficient evidence; both cases are unlikely to benefit from re-detection.

\subsection{Heuristic Scoring for Routing Priority}
To optimally route our limited high-resolution computational budget, the framework must accurately isolate specific regions of peak ambiguity across the heterogeneous spatial landscape. For a given multi-scale sliding window $w$, we compute the attention intensity $m(w) = \operatorname{mean}(\mathbf{H}_w)$ and the localized spatial attention entropy $\hat{H}_s(w)$. Rather than utilizing complex spatial intersection rules, we score each window by computing a joint \textit{ambiguity-saliency score}:
\begin{equation}
\sigma(w) = m(w) \cdot \hat{H}_s(w)
\end{equation}
This multiplicative formulation acts as a simple yet highly effective soft-AND gate for spatial routing. The term $m(w)$ functions as the semantic \textit{Saliency} indicator, confirming that the network detects the probable existence of an object and is actively attempting to route information to this spatial neighborhood. Conversely, $\hat{H}_s(w)$ quantifies the spatial \textit{Ambiguity}, indicating severe structural entanglement. Therefore, a high $\sigma(w)$ score effectively isolates a critical region: an area where potential target evidence exists ($m(w)$ is high) but is currently overwhelmed by extreme spatial confusion ($\hat{H}_s(w)$ is high). 

\subsection{Entropy-Guided Adaptive Spatial Routing}
Windows exhibiting a high ambiguity-saliency score represent highly suitable candidates for targeted computational resource allocation. We filter candidate windows using empirically validated thresholds $m(w) \ge \mu$ and $\hat{H}_s(w) \ge \tau_w$, apply class-agnostic non-maximum suppression (NMS) to eliminate spatial redundancy, and strictly select the top-$K$ highly ambiguous regions to bound the computational overhead.

For these specific, self-identified conflict zones, the framework executes \textit{entropy-guided adaptive spatial routing}. By extracting a high-resolution observation exclusively from the window $w$ and re-processing it, the system forcibly restricts the spatial prior. This dynamic routing mechanism directly injects localized, high-frequency visual evidence into the network's most ambiguous decision zones. By constraining the softmax normalization strictly to these localized features, the routing mechanism performs \textit{active variance reduction}. This instantly resolves the spatial ambiguity, facilitating the recovery of microscopic semantic entities without fundamentally altering the underlying base network weights.

\subsection{Multi-Scale Window Generation and Scoring}
We slide windows of relative scales $s\!\in\!\{0.25,0.5,0.75\}$ over $\mathbf{H}$ with stride $0.5$ of the window size.
We omit $s{=}1.0$ because the full image is already processed by the base detector; adding it would duplicate compute without adding spatial detail.
The minimum $64{\times}64$ crop avoids pathological tiny windows that degrade detector normalization and remove essential context for small objects.
For a window $w$ with top-left $(x,y)$, width $w_w{=}sW$ and height $w_h{=}sH$, the sub-heatmap is
\begin{equation}
\mathbf{H}_w=\mathbf{H}[y:y+w_h,\;x:x+w_w].
\end{equation}
We define the \emph{attention intensity}
\begin{equation}
m(w)=\operatorname{mean}(\mathbf{H}_w),
\end{equation}
normalize $\mathbf{H}_w$ to a distribution to obtain a window-level normalized entropy $\widehat H_s(w)\!\in\![0,1]$, and score each window by a soft-AND:
\begin{equation}
\label{eq:score}
\sigma(w)=m(w)\cdot \widehat H_s(w).
\end{equation}
We retain windows satisfying $m(w)\!\ge\!\mu$ and $\widehat H_s(w)\!\ge\!\tau_w$, apply class-agnostic non-maximum suppression (NMS, IoU$=0.5$), and keep the top-$K$ windows (default $K{=}5$), each with minimum size $64{\times}64$ pixels.

\paragraph{Why mean $\times$ entropy?}
$m(w)$ promotes regions already receiving non-trivial attention (presence of evidence), whereas $\widehat H_s(w)$ favors dispersed attention (ambiguity).
Their product focuses computation on regions that are \emph{both} informative and uncertain---precisely where higher resolution is most likely to recover fine details and correct confusions.

\paragraph{Alternative scoring variants.}
For completeness, we consider: \linebreak(i) mean-only $\sigma_{\text{mean}}(w){=}m(w)$; 
(ii) entropy-only $\sigma_{\text{ent}}(w){=}\widehat H_s(w)$; 
(iii) $\gamma$-sharpened mean $\sigma_{\gamma}(w){=}m(w)^{\gamma}\widehat H_s(w)$ with $\gamma{>}1$; and (iv) a soft-gated variant
\begin{equation}
\label{eq:soft_gate}
\begin{split}
\sigma_{\text{soft}}(w) &= \phi\!\bigl(m(w)\bigr)\,\widehat H_s(w),\\
\phi(t) &= \bigl(1+\exp[-k(t-\mu)]\bigr)^{-1},
\end{split}
\end{equation}
with $k{=}10$ by default.
Empirically, Eq.~\eqref{eq:score} provides a favorable recall--precision--cost balance.

\subsection{Crop Re-Detection and Fusion}
For each selected window $w{=}[x_1,y_1,x_2,y_2]$, we crop $\mathbf{I}_w$, resize to the detector's input size, and run the base detector to obtain refined detections $(\mathbf{b}_r,s_r,c_r)$.
For DETR-like models, normalized centers and sizes are converted to\linebreak
$(x_{\min}, y_{\min}, x_{\max}, y_{\max})$, scaled to the crop,
and translated by $(x_1, y_1)$ back to image coordinates.
We apply a mild entropy-aware score boost
\begin{equation}
\label{eq:boost}
s_r \leftarrow s_r\Bigl(1+0.2\bigl(\widehat H_s(w)-0.7\bigr)\Bigr),
\end{equation}
then merge refined detections with the initial full-image detections using class-wise non-maximum suppression (NMS, IoU$=0.5$) and a confidence threshold of $0.3$.
For overlapping boxes of the same class, we fuse scores by
\begin{equation}
\label{eq:fusion}
s_f=\alpha\, s_o + (1-\alpha)\, s_r,\qquad \alpha=0.7,
\end{equation}
keeping the box with the higher score.
This preserves precision of confident original detections while allowing refined crops to recover missed or under-confident tiny objects.

\subsection{Complexity and Runtime}
Let $C_{\text{det}}$ be the cost of one full-image forward pass.
ViCrop-Det adds: (i) attention extraction and heatmap construction, negligible relative to $C_{\text{det}}$ in modern implementations; and (ii) at most $K$ additional forwards on resized crops.
Let $r$ be the compute ratio between a crop and the full image ($r{<}1$).
The worst-case cost is thus
\begin{equation}
\label{eq:complexity}
C_{\text{total}}\approx \bigl(1+K\,r\bigr)\,C_{\text{det}}.
\end{equation}
The global gate in Eq.~\eqref{eq:gate} and the cap $K$ bound the overhead.
Empirically, many images either skip refinement or select $K'\!<\!K$ windows, yielding modest average slowdowns.
In addition, we report the mean number of selected windows $K'$ and the average crop-to-image compute ratio $\bar r$ per dataset in the appendix, which empirically confirms a modest overhead consistent with Eq.~\eqref{eq:complexity}.

\subsection{Implementation Details and Hyperparameters}
\label{subsec:impl-hparams}
Our goal is a \emph{single} robust configuration that transfers across detectors and datasets.
Hyperparameters were selected by a light grid search on the validation splits of DOTA-v1.5 and VisDrone, then \emph{frozen} for all results.

\begin{table}[ht]
\centering
\scriptsize
\caption{Default settings used for all models and datasets.}
\label{tab:hparams}
\resizebox{\linewidth}{!}{
\begin{tabular}{l l}
\toprule
\textbf{Component} & \textbf{Setting / Rationale} \\
\midrule
Attention aggregation & Mean over layers/heads/queries; stable across models. \\
Probability smoothing & $\varepsilon=10^{-8}$ in Eq.~\eqref{eq:prob_norm}. \\
SAE (global) & Eq.~\eqref{eq:sae}; $\widehat H_s\!\in[0,1]$. \\
Global gate & $\widehat H_s \ge \tau_g\ (=0.7)$ and $\operatorname{mean}(\mathbf{H}) \ge \mu\ (=0.3)$. \\
Scales / stride & $s\!\in\!\{0.25,0.5,0.75\}$; stride $0.5$. \\
Min window size & $64\times64$ px. \\
Window score & $\sigma(w)=m(w)\cdot \widehat H_s(w)$ (Eq.~\eqref{eq:score}). \\
Candidate thresholds & \begin{tabular}[t]{@{}l@{}} $m(w)\ge \mu\ (=0.3)$ and $\widehat H_s(w)\ge \tau_w\ (=0.7)$;\\ class-agnostic non-maximum suppression (NMS, IoU$=0.5$). \end{tabular} \\
Top-$K$ & $K=5$ (accuracy--efficiency trade-off). \\
Score boost & Eq.~\eqref{eq:boost}. \\
Score fusion & Eq.~\eqref{eq:fusion} with $\alpha=0.7$. \\
Final filtering & \begin{tabular}[t]{@{}l@{}} Class-wise non-maximum suppression (NMS, IoU$=0.5$); \\ confidence $\ge 0.3$. \end{tabular} \\
\bottomrule
\end{tabular}}
\end{table}

\paragraph{Design rationale.}
(1) \textit{Global gate.} $(\tau_g,\mu)=(0.7,0.3)$ suppresses images that are already well-focused or lack sufficient evidence.
(2) \textit{Multiplicative score.} $\sigma(w)=m\cdot\widehat H_s$ acts as a soft-AND; additive forms admitted more false positives in preliminary checks.
(3) \textit{Scales \& stride.} $\{0.25,0.5,0.75\}$ covers small-to-medium structures with manageable candidates; stride $0.5$ balances coverage and cost.
(4) \textit{Top-$K=5$.} Provides a good accuracy--efficiency trade-off; $K=3$ further reduces latency with a small AP drop.
(5) \textit{Fusion with $\alpha=0.7$.} Retains the precision of strong originals while allowing refined detections to correct misses.
(6) \textit{Averaging across layers/heads/queries.} Per-layer or per-head entropies are noisier and model-dependent; averaging yields a smoother, transferable spatial distribution that supports fixed thresholds with fewer false triggers.
\textbf{Across all tested DETR-style backbones and both datasets, the same $(\tau_g,\tau_w,\mu)$, $K{=}5$, and $\alpha{=}0.7$ remained effective without per-model retuning.}
\section{Experiment}

\begin{table*}[t]
\centering
\caption{Comparison on VisDrone and DOTA-v1.5.}
\setlength{\tabcolsep}{8pt}
\renewcommand{\arraystretch}{1.2}
\begin{adjustbox}{max width=\linewidth}
\begin{tabular}{llccc|ccc|ccc}
\toprule
\multirow{2}{*}{Dataset} & \multirow{2}{*}{Method} & \multicolumn{3}{c|}{RT-DETR-R50} & \multicolumn{3}{c|}{Deformable DETR} & \multicolumn{3}{c}{DINO-DETR} \\
\cmidrule(lr){3-5}\cmidrule(lr){6-8}\cmidrule(l){9-11}
 &  & mAP$_{50}$ & AP$_S$ & FPS & mAP$_{50}$ & AP$_S$ & FPS & mAP$_{50}$ & AP$_S$ & FPS \\
\midrule
\multirow{4}{*}{VisDrone}
& No            & 37.0 & 31.4 & 47.5 & 36.4 & 30.6 & 42.4 & 37.9 & 32.1 & 43.0 \\
& +ViCrop-Det   & \textbf{38.9} & \textbf{35.2} & 38.6 & \textbf{39.0} & \textbf{35.4} & 33.9 & \textbf{39.6} & \textbf{36.2} & 34.2 \\
& +SAHI         & 38.0 & 32.5 & 24.2 & 37.8 & 31.2 & 22.5 & 38.8 & 33.5 & 23.3 \\
& +ASAHI        & 38.2 & 34.3 & 27.5 & 38.1 & 33.9 & 26.7 & 39.4 & 35.8 & 26.5 \\
\midrule
\multirow{4}{*}{DOTA-v1.5}
& No            & 50.1 & 43.6 & 43.2 & 49.1 & 42.2 & 39.6 & 51.2 & 44.5 & 39.4 \\
& +ViCrop-Det   & \textbf{51.5} & \textbf{48.1} & 34.6 & \textbf{50.5} & \textbf{47.0} & 31.7 & \textbf{52.5} & \textbf{48.9} & 31.5 \\
& +SAHI         & 50.9 & 43.3 & 22.3 & 49.5 & 42.6 & 20.4 & 51.8 & 45.7 & 19.8 \\
& +ASAHI        & 51.2 & 45.0 & 28.5 & 49.8 & 44.5 & 26.6 & 52.1 & 46.6 & 26.3 \\
\bottomrule
\end{tabular}
\end{adjustbox}
\label{tab:method_comp_visdrone_dota}
\end{table*}

\begin{table}[t]
\small
\renewcommand{\arraystretch}{1.12}
\caption{MS COCO results with/without ViCrop--Det. AP$_S/_M/_L$ are computed on small ($a<32^2$), medium ($32^2\le a<96^2$), and large ($a\ge96^2$) objects, where $a$ is the GT box area (pixels$^2$).}
\label{tab:coco_vicropdet}
\resizebox{\linewidth}{!}{
\begin{tabular}{lcccc}
\toprule
\textbf{Metric} &
\multicolumn{2}{c}{\textbf{RT--DETR--R50}} &
\multicolumn{2}{c}{\textbf{Deformable DETR}} \\
& \textbf{no-ViCrop--Det} & \textbf{ViCrop--Det} & \textbf{no-ViCrop--Det} & \textbf{ViCrop--Det} \\
\midrule
mAP$_{50}$ & 71.3 & 72.1 & 62.6 & 63.8 \\
AP$_S$ & 34.8 & 36.9 & 26.4 & 29.0 \\
AP$_M$ & 58.0 & 58.2 & 47.1 & 47.4 \\
AP$_L$ & 70.0 & 70.0 & 58.0 & 58.0 \\
\bottomrule
\end{tabular}
}
\end{table}

\begin{table}[t]
\small
\centering
\caption{Ablation study on RT-DETR-R50 showing the impact of entropy-guided region selection versus random selection.}
\setlength{\tabcolsep}{8pt}
\begin{tabular}{lcc}
\toprule
\textbf{Dataset} & \textbf{Random-selection} & \textbf{Entropy-selection} \\
\midrule
VisDrone  & 37.5 & 38.9 \\
DOTA-v1.5 & 50.6 & 51.5 \\
\bottomrule
\end{tabular}
\label{tab:entropy_ablation}
\end{table}

\begin{table}[h]
\small
\centering
\caption{Varying $K$ on VisDrone. $K{=}5$ offers the best trade-off.}
\label{tab:K-ablation}
\setlength{\tabcolsep}{8pt}
\begin{tabular}{c|cccc}
\toprule
$K$         & 1 & 3 & 5 & 7 \\
\midrule
mAP$_{50}$  & 37.9 & 38.2 & \textbf{38.9} & 34.1 \\
FPS         & 42.5    & 40.2    & 38.6             & 31.5 \\
\bottomrule
\end{tabular}
\end{table}

\begin{figure*}[t]
  \centering
  \includegraphics[width=\textwidth]{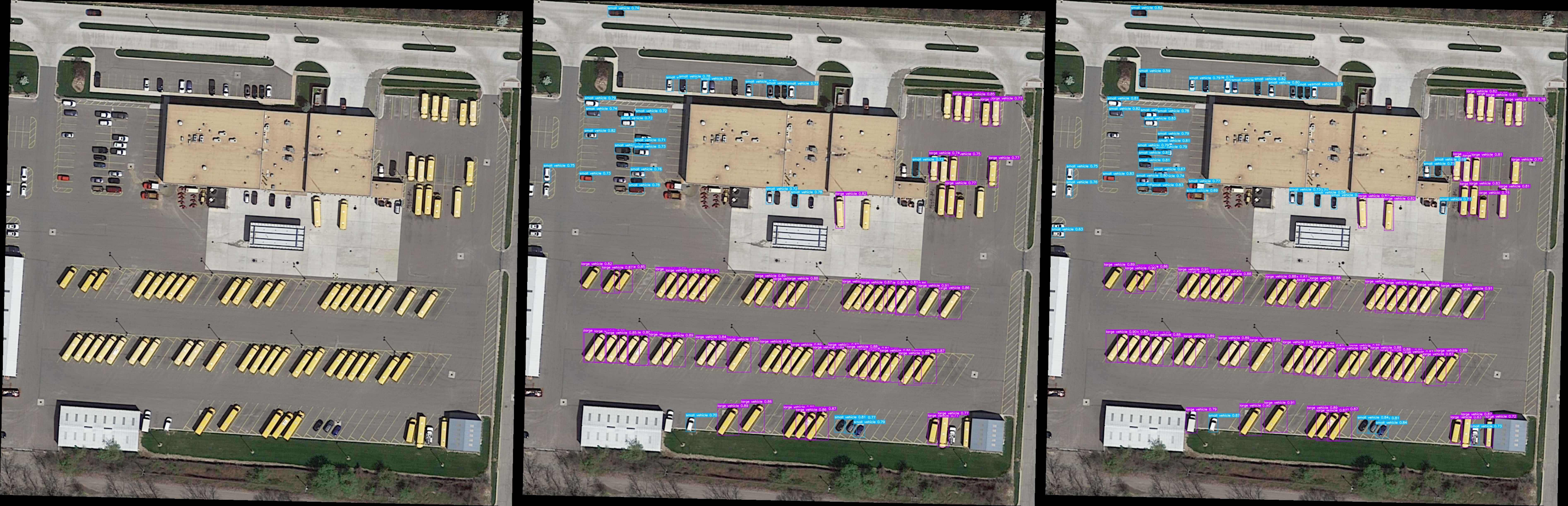}
  \caption{The RT-DETR-R50 model applies ViCrop-Det effect comparison diagram on the DOTA dataset (Original image; no-ViCrop-Det; ViCrop-Det).}
  \label{fig:rtdetr-dota}
\end{figure*}

\begin{figure*}[t]
  \centering
  \includegraphics[width=\textwidth]{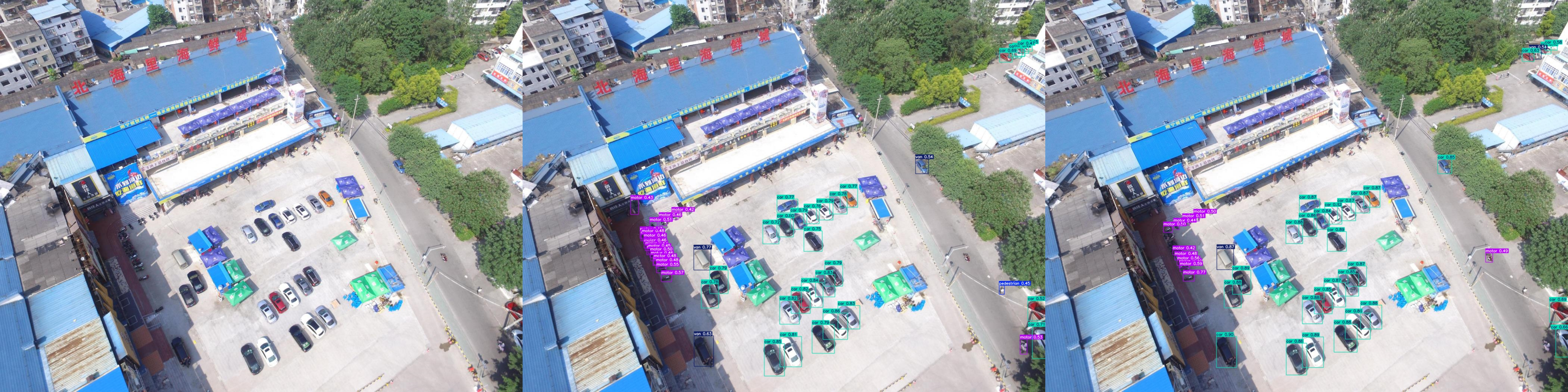}
  \caption{The RT-DETR-R50 model applies ViCrop-Det effect comparison diagram on the VisDrone dataset (Original image; no-ViCrop-Det; ViCrop-Det).}
  \label{fig:rtdetr-vis}
\end{figure*}

\begin{table}[h]
\small
\centering
\caption{Effect of entropy threshold on VisDrone. Best at $\tau{=}0.7$.}
\label{tab:entropy-ablation}
\setlength{\tabcolsep}{8pt}
\begin{tabular}{c|cccc}
\toprule
$\tau$ & 0.5 & 0.6 & 0.7 & 0.8 \\
\midrule
mAP$_{50}$  & 37.5 & 38.2 & \textbf{38.9} & 38.4 \\
AP$_{S}$    & 33.4 & 34.0 & \textbf{35.2} & 33.9 \\
\bottomrule
\end{tabular}
\end{table}

\subsection{Datasets}
\label{sec:datasets}

We evaluate ViCrop-Det on two challenging object detection benchmarks: DOTA-v1.5 and VisDrone-DET2019.

\textbf{DOTA-v1.5}~\cite{xia2018dota} is a large-scale remote sensing dataset comprising 2,806 high-resolution aerial images (ranging from $800\times800$ up to $20{,}000\times20{,}000$ pixels) captured by various airborne sensors. Compared to v1.0, v1.5 introduces exhaustive annotations for extremely small objects (smaller than 10 pixels) and adds a new category \textit{container crane}, resulting in a total of 16 object categories and 403,318 oriented bounding box instances. However, due to poor annotation quality of the newly added class, we exclude it in our experiments, effectively using 15 classes. The dataset is officially split into training, validation, and test subsets in a 1/2, 1/6, and 1/3 ratio, respectively.

\textbf{VisDrone-DET2019}~\cite{zhu2018visdrone} is a drone-view dataset captured over 14 cities in China under diverse lighting, weather, and altitude conditions. It includes 10,209 static images and 288 video sequences (261,908 frames) from UAV platforms. The object detection track provides axis-aligned bounding box annotations covering more than 2.6 million object instances across 10 categories. We use the image subset with an official split of 6,471/548/5,240 images for training, validation, and test-dev sets, respectively. An additional test-challenge set exists but is not used due to withheld annotations.

\textbf{MS COCO 2017}~\cite{lin2014microsoft} is a general-purpose detection benchmark with 80 everyday categories and diverse, cluttered scenes. We follow the standard splits, using \textit{train2017} ($\approx$118k images) and \textit{val2017} (5k) with the official COCO evaluation protocol to assess the cross-domain generalization of \emph{ViCrop--Det}. Note that COCO is not a primary evaluation dataset in this paper; it is mainly used to verify that our inference framework improves small-object detection without degrading the base model's performance on medium/large objects.
\subsection{Comparative Experiments}
\label{subsec:comparative}

\paragraph{Overall comparison.}
Across all three backbones (RT--DETR--R50, Deformable DETR, DINO--DETR) and both benchmarks (VisDrone, DOTA-v1.5), \textbf{ViCrop--Det} offers the strongest accuracy--throughput trade-off (Table~\ref{tab:method_comp_visdrone_dota}). 
On \textbf{VisDrone}, mAP$_{50}$ increases by \mbox{+1.7--2.6}~pp and AP$_S$ by \mbox{+3.8--4.8}~pp, while sustaining \mbox{\textbf{33.9--38.6 FPS}} (vs.\ 42.4--47.5 baseline). 
On \textbf{DOTA-v1.5}, gains are \mbox{+1.3--1.4}~pp mAP$_{50}$ and \mbox{+4.4--4.8}~pp AP$_S$ at \mbox{\textbf{31.5--34.6 FPS}} (vs.\ 39.4--43.2 baseline). 
Notably, the improvements are dominated by AP$_S$, indicating that routing computation to high-attention, high-entropy regions preferentially recovers small objects while preserving overall detection quality. 
Per-class results are provided in the supplementary.

\paragraph{Against slicing baselines.}
Under the same backbone, \textbf{SAHI/ASAHI} deliver smaller accuracy gains with substantially lower throughput.
For example on \textbf{VisDrone} (RT--DETR--R50), SAHI reaches 38.0/32.5 at \textbf{24.2 FPS} and ASAHI 38.2/34.3 at \textbf{27.5 FPS}, whereas \textbf{ViCrop--Det} achieves \textbf{38.9/35.2 at 38.6 FPS}. 
Similar gaps hold for Deformable DETR and DINO--DETR as well as \textbf{DOTA-v1.5}: ViCrop--Det is consistently \mbox{1.47--1.60$\times$} faster than SAHI and \mbox{1.19--1.40$\times$} faster than ASAHI, while also yielding higher AP$_S$ (\mbox{+2.7--4.8}~pp vs.\ SAHI; \mbox{+0.4--3.1}~pp vs.\ ASAHI). 
This validates that entropy-guided, selective refinement is more efficient than uniform or coarse adaptive slicing.

\paragraph{Efficiency Overhead.}
ViCrop-Det acts purely as an inference--time post--processing module.  Let
\(C_{\mathrm{det}}\) denote the computational cost of a single full-image
forward pass.  In the worst case the additional cost introduced by
ViCrop-Det is bounded by \(K\,r\,C_{\mathrm{det}}\), where \(K=5\) is the
maximum number of high-entropy windows re-inspected per image and
\(r{<}1\) is the crop--to--image cost ratio
(\(r\!\approx\!0.25\) in our implementation).  Because most images
either \emph{(i)} select fewer than five windows or \emph{(ii)} are
skipped entirely by the entropy gate, the observed throughput drop is
only about \textbf{20--25\%} in practice.  A complete timing breakdown
for each dataset is provided in the supplementary material.

\paragraph{Generalization experiment}
To validate the generality of our method, we evaluate ViCrop-Det on the large-scale and diverse MS COCO dataset, which contains a wide distribution of object sizes across 80 categories. As shown in Table~\ref{tab:coco_vicropdet}, ViCrop-Det consistently improves the mAP@50 for both RT-DETR-R50 and Deformable DETR baselines. Notably, AP$_S$---which reflects performance on small objects ($a{<}32^2$ pixels)---improves significantly: from 34.8\% to 36.9\% on RT-DETR-R50 and from 26.4\% to 29.0\% on Deformable DETR. Meanwhile, performance on medium and large objects (AP$_M$, AP$_L$) remains unchanged or slightly improved, indicating that our refinement selectively enhances detection of small, ambiguous regions without degrading the detector's original strengths on larger objects. These results confirm the universality of ViCrop-Det as a plug-and-play test-time enhancement for Transformer detectors across domains and object scales.

\paragraph{Take-Away.}
ViCrop-Det offers a training-free, model-agnostic route to stronger
small-object detection: one pass of the base detector, followed by
entropy-guided high-resolution refinement, yields uniform per-class
improvements---especially on tiny or long-tail categories---with
manageable compute and zero architectural changes.

\subsection{Ablation experiment}
\label{subsec:ablation}
\paragraph{Entropy-Guided Window Selection (Equal-Budget Ablation)}
Crucially, both variants use exactly the same crop budget (identical $K$,
window scales and stride), ensuring that improvements stem from \emph{where}
computation is routed rather than \emph{how much} computation is used.
A key hypothesis of \textbf{ViCrop-Det} is that the \emph{spatial attention entropy} computed from the aggregated decoder cross-attention distribution, \emph{when coupled with attention intensity}, is an effective proxy for routing limited high-resolution computation to regions that are both salient and ambiguous. Concretely, all our ablations in this section use the window score \mbox{$\sigma(w)=m(w)\cdot\widehat H_s(w)$}, where $m(w)$ is the mean attention in the window and $\widehat H_s(w)$ is the normalized spatial attention entropy.

To isolate the effect of entropy-guided selection, we compare our \textit{Entropy-selection} against a \textit{Random-selection} baseline that samples the same number of windows uniformly at random while keeping \emph{every other component identical}: window scales ${0.25,0.5,0.75}$, stride $0.5$, thresholds $m\!\ge\!0.3$ and $\widehat H_s\!\ge\!0.7$, class-agnostic NMS with IoU$=0.5$, Top-$K=5$, confidence threshold $0.3$, and score fusion $s_f=\alpha s_o+(1-\alpha)s_r$ with $\alpha=0.7$.

On \textbf{VisDrone}, entropy guidance yields a \textbf{+1.4 pp} absolute gain, and on the more challenging \textbf{DOTA-v1.5} it delivers \textbf{+0.9 pp} (Table~\ref{tab:entropy_ablation}). These improvements are obtained \emph{without} increasing the number of crops or the inference budget, indicating that high-entropy, high-attention regions concentrate the detector's ambiguity more effectively than randomly chosen windows. In contrast, random cropping spends additional passes on background or already-confident areas, diluting gains. Based on these results, we adopt entropy-guided selection in all subsequent experiments.

\paragraph{Effect of the number of crops ($K$).}
As shown in Table~\ref{tab:K-ablation}, increasing $K$ monotonically reduces throughput (38/33/30/24 FPS for $K{=}1/3/5/7$), matching the near-linear compute overhead of additional forward passes. Accuracy improves from $K{=}1$ to $K{=}5$ (mAP${50}$: 37.9 $\rightarrow$ 38.2 $\rightarrow$ \textbf{38.9}), indicating that multiple, targeted refinements are needed to recover fine details in cluttered scenes. Pushing to $K{=}7$ degrades performance (34.1 mAP${50}$), suggesting diminishing returns and over-refinement: more overlapping crops introduce redundant hypotheses, exacerbate NMS/fusion conflicts, and allocate budget to lower-entropy regions that amplify false positives or score dilution. We therefore adopt $K{=}5$ as the default, which yields the best accuracy--efficiency trade-off (38.6 FPS). For latency-sensitive deployments, $K{=}3$ provides a reasonable alternative (40.2 FPS with minor accuracy loss). Note that the global gate often skips easy images, so the realized average number of crops $K'$ is typically below the upper bound $K$, further reducing the amortized overhead.

\paragraph{Effect of the Entropy Threshold $\tau$}
\label{subsec:tau_ablation}
We ablate the per\mbox{-}window entropy threshold $\tau$ used to filter candidate crops (Sec.~\ref{sec:method}), while keeping all other components fixed: the attention\mbox{-}intensity threshold $m(w)\!\ge\!0.3$, window scales $\{0.25,0.5,0.75\}$ with stride $0.5$, top-$K\!=\!5$, class\mbox{-}agnostic NMS (IoU$=0.5$), and the fusion strategy. As shown in Table~\ref{tab:entropy-ablation}, performance follows a clear concave trend and \textbf{peaks at $\tau{=}0.7$}: mAP$_{50}$ improves from $37.5$ ($\tau{=}0.5$) to \textbf{38.9} (+1.4~pp), and AP$_S$ rises from $33.4$ to \textbf{35.2} (+1.8~pp). 

We attribute this behavior to the role of $\tau$ in balancing \emph{coverage} and \emph{selectivity}. A low threshold (e.g., $\tau{=}0.5$) is overly permissive, admitting many low\mbox{-}uncertainty windows; this spreads the crop budget onto background or already confident regions, diluting small\mbox{-}object gains. Conversely, a high threshold (e.g., $\tau{=}0.8$) becomes too strict, suppressing ambiguous yet informative windows; accuracy drops to 38.4 mAP$_{50}$ and 33.9 AP$_S$. The sweet spot at \textbf{$\tau{=}0.7$} best aligns with our routing principle---allocating limited high\mbox{-}resolution compute to regions that are \emph{both} salient (high attention) and ambiguous (high entropy). We therefore \textbf{fix $\tau{=}0.7$} for all subsequent experiments.

\subsection{Visual analysis}

\noindent\textbf{Setup and visuals.}
Fig.~\ref{fig:rtdetr-dota} and Fig.~\ref{fig:rtdetr-vis} juxtapose detections from the base RT--DETR--R50 against the same model augmented with \textbf{ViCrop-Det} on \textit{DOTA-v1.5} and \textit{VisDrone}, respectively, revealing clearer boundaries, fewer false positives, and improved recall on tiny, cluttered targets. 

\noindent\textbf{DOTA-v1.5: structural complexity under clutter.}
On high-resolution aerial scenes, ViCrop-Det consistently strengthens categories that are structurally complex or appear at small scales. For RT--DETR--R50, \texttt{BRIDGE} improves \mbox{14.2$\rightarrow$16.3} (+2.1), \texttt{ROUNDABOUT} \mbox{15.5$\rightarrow$18.1} (+2.6), and \texttt{SOCCER BALL FIELD} \mbox{36.1$\rightarrow$38.2} (+2.1); strong classes such as \texttt{PLANE}, \texttt{LARGE VEHICLE}, and \texttt{TENNIS COURT} retain or slightly increase accuracy. The overall mAP50 rises from \mbox{0.501$\rightarrow$51.5}. 

\noindent\textbf{VisDrone: gains on small and long-tail classes.}
In dense UAV imagery, the largest absolute gains concentrate on low-baseline or long-tail categories. With RT--DETR--R50, \texttt{BICYCLE} improves \mbox{12.5$\rightarrow$17.2} (+4.7), \texttt{AWNING-TRICYCLE} \mbox{8.8$\rightarrow$12.9} (+4.1), \texttt{TRICYCLE} \mbox{21.3$\rightarrow$25.4} (+4.1), and \texttt{PEOPLE} \mbox{30.3$\rightarrow$31.2}; the overall mAP50 increases \mbox{37.0$\rightarrow$38.9}. Deformable~DETR exhibits the same trend, e.g., overall \mbox{36.4$\rightarrow$39.0}. 

\noindent\textbf{Uniform, class-wise improvements.}
A salient property is universality: across two detectors and two datasets, \emph{every} reported class improves after integrating ViCrop-Det, ruling out the possibility that overall gains are driven by a few dominant categories. For fairness on DOTA-v1.5, the low-quality \texttt{Container~Crane} category is excluded, and the gains cover the remaining 15 classes.

\noindent\textbf{Why the visuals improve.}
ViCrop-Det extracts decoder cross-attention, computes a normalized \emph{spatial attention entropy}, and scores each multi-scale window by $\sigma = \text{mean}\times\text{entropy}$ to prioritize regions that are both highly attended and highly uncertain; only the top-$K$ windows (default $K{=}5$) are re-detected at high resolution. This selective routing prevents wasted computation on background, and the final fusion uses class-wise NMS (IoU~0.5) with a confidence threshold of 0.3 and a weighted score merge $s_f=\alpha s_o + (1-\alpha)s_r$ ($\alpha{=}0.7$), coupled with a mild entropy-aware score boost for refined boxes.

\section{Limitations}
ViCrop-Det relies on fixed thresholds, which may not adapt per-dataset, and misses zero-attention objects (future: integrate saliency). Compute overhead (20-25\% throughput drop) is bounded but higher for dense scenes. Broader datasets reveal less gain on large objects, emphasizing small-object focus.
\section{Conclusion}
In this paper, we introduced ViCrop-Det, a novel training-free inference framework designed to enhance small object detection in Transformer-based architectures. Moving away from rigid uniform slicing strategies, we demonstrated that the decoder's endogenous cross-attention distribution can serve as a powerful tool for localized refinement. By proposing a joint ambiguity-saliency heuristic scoring mechanism based on Spatial Attention Entropy (SAE), our method effectively isolates dense, conflict-prone regions. This allows ViCrop-Det to perform targeted, high-resolution adaptive spatial routing exclusively on areas suffering from severe cognitive ambiguity, without requiring model retraining or architectural modifications. 

Experimental evaluations on challenging benchmarks like VisDrone and DOTA-v1.5 demonstrate that this entropy-guided approach consistently improves performance across diverse baseline detectors, including RT-DETR-R50 and Deformable DETR. It yields significant mAP gains particularly for small and heavily occluded objects, achieving a highly optimized accuracy-efficiency trade-off compared to conventional brute-force slicing methods. While ViCrop-Det introduces a bounded computational overhead and inherently relies on the base detector's initial attention quality, its plug-and-play versatility highlights a highly pragmatic direction for deploying robust perception models in high-resolution, cluttered real-world environments. Future work will explore dynamic thresholding mechanisms for domain-specific optimization and extensions to video-level spatial-temporal routing.

{
  \small
  \bibliographystyle{unsrtnat}
  \bibliography{main}
}

\clearpage
\appendix
\section*{Supplementary Material}
\addcontentsline{toc}{section}{Supplementary Material}
\setcounter{figure}{0}
\setcounter{table}{0}
\setcounter{equation}{0}
\renewcommand{\thefigure}{S\arabic{figure}}
\renewcommand{\thetable}{S\arabic{table}}
\renewcommand{\theequation}{S\arabic{equation}}

\begin{abstract}
This supplementary material provides additional details on the experimental settings, the throughput measurement protocol, runtime overhead analysis, extended ablation studies, and supplementary explanation of the method section presented in the main paper.
\end{abstract}


\section{Experimental Settings}

\paragraph{Hardware/Software.} All experiments were performed using an RTX~4090 (24GB), an i7--13700KF processor, and 32GB DDR5@6400 memory. The software stack includes Python~3.10.12, PyTorch~2.0.1 (CUDA~11.8), cuDNN~8.9, NCCL~2.17, and OpenCV~4.8. 

\paragraph{Datasets.} We evaluate ViCrop-Det across four object detection benchmarks:

\textbf{DOTA-v1.5} (2,806 images, 16 classes, 403,318 instances)

\textbf{VisDrone-DET2019} (10,209 images, 10 classes, over 2.6M boxes)

\textbf{VOC 2012} (20,000 images, 20 classes, 11,530 instances)

\textbf{MS COCO} (123,287 images, 80 classes, over 2.5M instances)

The datasets are used with their official splits, and we report results on the validation sets unless otherwise specified. The datasets were pre-processed using standard augmentation techniques, such as random flipping, scaling, and color jittering, to ensure robustness across different conditions.

\paragraph{Hyperparameters (frozen).} The key hyperparameters used in our experiments are:

Entropy threshold: $0.7$

Mean-attention threshold: $0.3$

Scales for cropping: ${0.25, 0.5, 0.75}$

Stride for windowing: $0.5$

Crop budget: $K = 5$

Minimum crop size: $64$ px

NMS IoU threshold for crop selection: $0.5$

Confidence threshold: $0.3$

Fusion weight: $\alpha = 0.7$

The hyperparameters were selected through a light grid search on the validation splits of DOTA-v1.5 and VisDrone. The chosen values strike a balance between detection accuracy and computational efficiency for small object detection tasks.

\section{Throughput Measurement Protocol}
We measure throughput using end-to-end wall-clock time on validation images. The batch size is set to $B = 1$, and the first $W = 20$ iterations are considered warm-up, with the next $N = 200$ iterations being timed. The protocol includes synchronization before and after each forward pass to ensure accurate timing.

\begin{algorithm}[h!]
\caption{FPS Benchmark Routine (End-to-End Mode)}
\label{alg:fps_bench}
\begin{algorithmic}[1]
\For{$j \gets 1$ \textbf{to} $W$}
\State $\mathcal{M}(\mathcal{I}_j)$ \Comment{warm-up, untimed}
\EndFor
\State $t \gets 0$
\For{$i \gets 1$ \textbf{to} $N$}
\State \texttt{cuda.synchronize()}
\State $s \gets$ \texttt{time.time()}
\State $\mathcal{M}(\mathcal{I}_i)$
\State \texttt{cuda.synchronize()}
\State $t \gets t + (\texttt{time.time()} - s)$
\EndFor
\State \textbf{return} $(\mathrm{FPS}=N/t,\; \mathrm{ms/img}=1000/\mathrm{FPS})$
\end{algorithmic}
\end{algorithm}

\paragraph{Efficiency Optimizations.} Throughput was measured across various batch sizes, and we optimized the code for single-image processing to minimize the impact of batch size on throughput. ViCrop-Det achieves a significant speedup over traditional slicing methods, as seen in the comparison to SAHI and ASAHI. The procedure outlined in Algorithm~\ref{alg:fps_bench} ensures that throughput is measured accurately by excluding warm-up iterations and focusing on the final 200 iterations for timing.

\section{Runtime Overhead Analysis}
On RT--DETR--R50, the ViCrop-Det framework reduces throughput by approximately $20$--$23\%$, while still maintaining real-time speed, validating its practicality in real-world applications. This overhead is minimized through selective refinement based on entropy, which avoids unnecessary computation on already well-focused regions.

\section{Extended Comparative Experiment}

\begin{table}[t]
\centering
\caption{Comparison on VOC 2012.}
\setlength{\tabcolsep}{8pt}
\renewcommand{\arraystretch}{1.2}
\begin{adjustbox}{max width=\linewidth}
\begin{tabular}{llccc|ccc}
\toprule
\multirow{2}{*}{Dataset} & \multirow{2}{*}{Method} & \multicolumn{3}{c}{RT-DETR-R50} & \multicolumn{3}{c}{Deformable DETR} \\
\cmidrule(lr){3-5}\cmidrule(l){6-8}
 &  & mAP$_{50}$ & AP$_S$ & FPS & mAP$_{50}$ & AP$_S$ & FPS \\
\midrule
\multirow{4}{*}{VOC 2012}
& No            & 81.6 & 75.4 & 84.5 & 79.7 & 72.3 & 81.2 \\
& +ViCrop-Det   & \textbf{82.1} & \textbf{77.6} & 67.2 & \textbf{80.1} & \textbf{74.8} & 65.3 \\
& +SAHI         & 81.8 & 75.5 & 43.2 & 79.8 & 72.5 & 42.0 \\
& +ASAHI        & 81.9 & 76.1 & 47.3 & 79.8 & 72.6 & 44.1 \\
\bottomrule
\end{tabular}
\end{adjustbox}
\label{tab:comp_voc2012}
\end{table}

The results from Table~\ref{tab:comp_voc2012} highlight that ViCrop-Det consistently outperforms both SAHI and ASAHI in terms of small-object detection (AP$_S$), showing a significant improvement in the detection of small and hard-to-detect objects. While ViCrop-Det introduces some computational overhead, as evidenced by the drop in FPS, it provides a clear advantage in terms of detection accuracy, particularly for small-object categories.

The slicing methods, while offering some gains in small-object detection, fall short in terms of both detection precision (AP$_S$) and runtime efficiency, making ViCrop-Det a more efficient and effective solution for improving small-object detection in real-time applications. The modest increase in runtime for ViCrop-Det is justified by its enhanced detection performance, particularly in the context of applications where small-object detection is crucial, such as surveillance, robotics, and autonomous driving.

\section{Extended Ablation Studies}
We provide additional ablation studies conducted using the VisDrone validation set, leveraging the RT--DETR--R50 model.

\subsection{Confidence Score Fusion}
\begin{table}[h]
\small
\centering
\caption{Effect of confidence score fusion on VisDrone.}
\label{tab:score-fusion}
\setlength{\tabcolsep}{10pt}
\begin{tabular}{l|c}
\toprule
Strategy & mAP$_{50}$ \\
\midrule
Without fusion & 38.6 \\
With fusion & \textbf{38.9} \\
\bottomrule
\end{tabular}
\end{table}

In this ablation study, we investigate the effect of confidence score fusion on the detection performance of ViCrop-Det using the VisDrone dataset. The primary goal of confidence score fusion is to combine the original detections from the model with the refined predictions obtained through high-resolution re-detection of ambiguous regions, as guided by the spatial attention entropy (SAE).

In Table~\ref{tab:score-fusion}, we compare the performance of the model with and without the confidence score fusion. The results show that the inclusion of confidence score fusion leads to a slight improvement in the mean average precision at 50\% Intersection over Union (mAP$_{50}$), from 38.6 to 38.9, representing an absolute gain of 0.3 points.

This improvement is indicative of the benefit of fusing refined predictions with the original detections. The refined predictions focus on the high-entropy, ambiguous regions of the image where the model's confidence is lower, particularly for small or cluttered objects. By incorporating these refined detections, the model can recover missed or under-confident small objects, thus boosting overall detection accuracy.

The increase in mAP$_{50}$ is modest, yet significant, highlighting that the method does not introduce substantial noise or false positives but instead refines the model's performance in challenging regions, particularly small and occluded objects. This confirms that confidence score fusion is an effective strategy for enhancing the precision of small-object detection in ViCrop-Det without causing degradation in the overall performance.

\subsection{Per-Class Results on VisDrone and DOTA-v1.5}
Tables~\ref{tab:visdrone_merge} and~\ref{tab:dota_merge} provide a detailed comparison of per-class $\mathrm{mAP}_{50}$ scores with and without ViCrop-Det. Results show significant improvements across both RT--DETR--R50 and Deformable DETR on both datasets, particularly for small object categories.

\begin{table*}[t]
\small
\centering
\caption{Per-class $\mathrm{mAP}_{50}$ on \textbf{VisDrone} with and without \textbf{ViCrop-Det}, comparing \textbf{RT--DETR--R50} and \textbf{Deformable DETR}. Parentheses show absolute improvement (ViCrop-Det $-$ no-ViCrop-Det).}
\label{tab:visdrone_merge}
\setlength{\tabcolsep}{6pt}
\renewcommand{\arraystretch}{1.05}
\begin{tabular}{lcccc}
\toprule
\multirow{2}{*}{\textbf{Class}} &
\multicolumn{2}{c}{\textbf{RT--DETR--R50}} &
\multicolumn{2}{c}{\textbf{Deformable DETR}} \\
\cmidrule(lr){2-3}\cmidrule(lr){4-5}
& \textbf{no-ViCrop-Det} & \textbf{ViCrop-Det} & \textbf{no-ViCrop-Det} & \textbf{ViCrop-Det} \\
\midrule
ALL & 37.0 & 38.9 (+1.9) & 36.4 & 39.0 (+2.6) \\
PEDESTRIAN & 41.6 & 42.5 (+0.9) & 40.4 & 42.6 (+2.2) \\
PEOPLE & 30.3 & 31.2 (+0.9) & 30.6 & 34.7 (+4.1) \\
BICYCLE & 12.5 & 17.2 (+4.7) & 11.9 & 15.4 (+3.5) \\
CAR & 79.7 & 80.3 (+0.6) & 79.4 & 82.1 (+2.7) \\
VAN & 43.7 & 43.9 (+0.2) & 43.2 & 45.8 (+2.6) \\
TRUCK & 36.0 & 38.1 (+2.1) & 37.1 & 39.6 (+2.5) \\
TRICYCLE & 21.3 & 25.4 (+4.1) & 20.4 & 23.5 (+3.1) \\
AWNING-TRICYCLE & 8.8 & 12.9 (+4.1) & 9.2 & 10.9 (+1.7) \\
BUS & 52.6 & 53.4 (+0.8) & 50.3 & 51.7 (+1.4) \\
MOTOR & 43.1 & 43.6 (+0.5) & 41.2 & 43.3 (+2.1) \\
\bottomrule
\end{tabular}
\end{table*}

\begin{table*}[t]
\small
\centering
\caption{Per-class $\mathrm{mAP}_{50}$ on \textit{DOTA-v1.5} with and without \textbf{ViCrop-Det}, comparing \textbf{RT--DETR--R50} and \textbf{Deformable DETR}. Parentheses show absolute improvement (ViCrop-Det $-$ no-ViCrop-Det).}
\label{tab:dota_merge}
\setlength{\tabcolsep}{6pt}
\renewcommand{\arraystretch}{1.05}
\begin{tabular}{lcccc}
\toprule
\multirow{2}{*}{\textbf{Class}} &
\multicolumn{2}{c}{\textbf{RT--DETR--R50}} &
\multicolumn{2}{c}{\textbf{Deformable DETR}} \\
\cmidrule(lr){2-3}\cmidrule(lr){4-5}
& \textbf{no-ViCrop-Det} & \textbf{ViCrop-Det} & \textbf{no-ViCrop-Det} & \textbf{ViCrop-Det} \\
\midrule
ALL & 50.1 & 51.5 (+1.4) & 49.1 & 50.5 (+1.4) \\
PLANE & 74.0 & 75.2 (+1.2) & 72.1 & 73.5 (+1.4) \\
SHIP & 61.4 & 62.4 (+1.0) & 56.5 & 58.2 (+1.7) \\
STORAGE TANK & 34.7 & 36.1 (+1.4) & 31.4 & 34.1 (+2.7) \\
BASEBALL DIAMOND & 52.3 & 54.4 (+2.1) & 51.8 & 52.5 (+0.7) \\
TENNIS COURT & 92.3 & 92.5 (+0.2) & 89.9 & 91.2 (+1.3) \\
BASKETBALL COURT & 38.6 & 39.2 (+0.6) & 38.2 & 39.4 (+1.2) \\
GROUND TRACK FIELD & 35.1 & 36.8 (+1.7) & 37.1 & 37.8 (+0.7) \\
HARBOR & 72.5 & 74.1 (+1.6) & 71.6 & 72.1 (+0.5) \\
BRIDGE & 14.2 & 16.3 (+2.1) & 16.2 & 18.5 (+2.3) \\
LARGE VEHICLE & 79.8 & 80.6 (+0.8) & 78.4 & 79.3 (+0.9) \\
SMALL VEHICLE & 57.2 & 58.5 (+1.3) & 55.0 & 56.2 (+1.2) \\
HELICOPTER & 42.7 & 43.5 (+0.8) & 43.5 & 45.0 (+1.5) \\
ROUNDABOUT & 15.5 & 18.1 (+2.6) & 14.7 & 16.3 (+1.6) \\
SOCCER BALL FIELD & 36.1 & 38.2 (+2.1) & 35.4 & 38.5 (+3.1) \\
SWIMMING POOL & 45.6 & 47.0 (+1.4) & 44.6 & 45.9 (+1.3) \\
\bottomrule
\end{tabular}
\end{table*}

\section{Additional Explanation of the Method Section}

In the main paper, we introduced ViCrop-Det, a training-free, plug-and-play framework designed to improve the detection of small objects in Transformer-based detectors, such as RT--DETR--R50 and Deformable DETR. Here, we provide a deeper explanation of key components of the method, including the motivation behind spatial attention entropy (SAE), the selection of ambiguous regions, and the efficiency of the multi-scale cropping process.

\subsection{Spatial Attention Entropy (SAE)}

The core idea behind ViCrop-Det is to leverage internal decoder cross-attention statistics from a single forward pass to guide the detection refinement process. By aggregating cross-attention from multiple layers and heads, we generate a high-resolution spatial attention map. From this map, we compute the Shannon entropy (SAE), which serves as an uncertainty measure for the regions in the image. High entropy regions correspond to areas of ambiguity, where the model is uncertain and likely to benefit from higher-resolution reprocessing. This entropy-guided mechanism ensures that computation is directed toward the most uncertain yet potentially informative regions.

In Transformer-based detectors, the attention mechanism computes interactions between the query tokens and key tokens, allowing the model to focus on relevant regions in the image. For \textbf{ViCrop-Det}, we aggregate the cross-attention maps produced by multiple layers and heads in the decoder. This aggregation provides a high-resolution \textbf{spatial attention map}, which captures the focus of the model across the image, highlighting areas of varying importance during the detection process.

To better understand the attention map, we average across the layers, heads, and queries, reducing noise and retaining the most significant information about the model's focus. This aggregated attention map indicates where the model places its most significant focus. Importantly, areas with high attention intensity reflect regions where the model believes there is strong evidence for detection, while regions with low attention intensity might indicate less confident detections.

\paragraph{Computing Spatial Attention Entropy (SAE)}
Once we have the aggregated spatial attention map, we compute the \textbf{Shannon entropy (SAE)}, which quantifies the uncertainty in the model's attention distribution. The entropy is calculated as follows:

\begin{equation}
H_s = - \sum_{i \in P} p_i \log p_i
\end{equation}
where \( p_i \) is the normalized attention score for the \( i \)-th spatial location. \textbf{High entropy values} indicate regions where the model's attention is spread out, meaning it is uncertain about which part of the image to focus on. On the other hand, \textbf{low entropy values} signify focused attention, where the model is confident about the regions it is attending to.

The entropy map provides a clear indication of the areas where the model's focus is most ambiguous and could potentially benefit from higher resolution or reprocessing. High entropy regions are typically small or occluded objects, which are often harder to detect due to issues such as scale reduction, occlusions, or background clutter. The computation of SAE enables us to pinpoint these regions, allowing for targeted re-detection and refinement.

\paragraph{Role of SAE in Guiding Computation}
The key benefit of using \textbf{SAE} is that it provides an \textbf{uncertainty measure} for different regions in the image. By computing the entropy of the attention map, we can effectively direct computational resources to the areas that need it most, specifically the regions where the model is uncertain. These high-entropy areas often correspond to \textbf{small objects}, \textbf{occluded objects}, or \textbf{ambiguous regions} where attention is diluted due to overlapping background features.

The primary role of SAE is to guide the computational budget of \textbf{ViCrop-Det} to regions that are more likely to benefit from higher resolution processing. These areas are typically the ones that require more detailed analysis, such as small objects that were downsampled during the initial processing or regions that overlap with background clutter, causing misclassification. The selective refinement based on SAE ensures that \textbf{ViCrop-Det} focuses computational power on \textbf{uncertain regions} where the model is likely to improve its detection performance.

In practice, regions with \textbf{high entropy} are the ones where the model needs further refinement. By applying multi-scale cropping to these areas, we can perform high-resolution re-detection, enabling the model to recover fine details that are otherwise lost due to downsampling or attention dilution. This targeted refinement strategy enhances the model's ability to detect \textbf{small objects} and \textbf{ambiguous cases}, improving detection accuracy while minimizing unnecessary computations.

\paragraph{Benefits of SAE-Based Refinement}
The primary advantage of using SAE to guide the refinement process is that it ensures computational efficiency. Instead of performing exhaustive reprocessing over the entire image, \textbf{ViCrop-Det} only focuses on high-entropy regions, reducing the computational cost while still improving detection accuracy. This selective attention to \textbf{uncertain areas} leads to significant performance improvements, particularly in the detection of small or cluttered objects.

Moreover, by selectively refining high-entropy regions, \textbf{ViCrop-Det} maintains a balance between \textbf{accuracy} and \textbf{efficiency}. The model achieves better performance in detecting small and ambiguous objects without wasting computational resources on well-detected, low-entropy regions. This makes \textbf{ViCrop-Det} suitable for real-time applications, where computational overhead must be minimized.

\subsection{Targeted Region Selection and Cropping}

Using the computed Spatial Attention Entropy (SAE), we prioritize regions that exhibit both high attention intensity and high entropy for further refinement. This dual focus ensures that computational resources are applied to areas where the model is uncertain, particularly small or occluded objects, and where high-resolution reprocessing is most beneficial.

\paragraph{Multi-Scale Sliding Windows}
We apply multi-scale sliding windows at scales ${0.25, 0.5, 0.75}$ to focus on regions of different sizes in the image. These scales are chosen to capture a broad range of object sizes, from very small objects (scale $0.25$) to medium-sized objects (scale $0.75$). This multi-scale approach is critical for addressing the challenge of small-object detection, where the size of the object significantly impacts its detectability.

The sliding windows move across the attention map with a \textbf{stride of $0.5$}, meaning that the windows overlap with each other by 50

\paragraph{Scoring Mechanism}
Each window is scored based on a multiplicative combination of \textbf{attention intensity} and \textbf{SAE}. The attention intensity \( m(w) \) is computed as the \textbf{mean attention} across the pixels within the window \( w \), and SAE \( H_s(w) \) is the spatial entropy within the window. This score, defined as:

\begin{equation}
\sigma(w) = m(w) \times H_s(w)
\end{equation}

combines two critical components:
1. \textbf{Attention Intensity (\( m(w) \))}: This quantifies the strength of the model's focus on the window. A higher mean attention indicates that the model is already placing significant emphasis on the region.
2. \textbf{Spatial Attention Entropy (\( H_s(w) \))}: This quantifies the uncertainty in the model's attention. High entropy indicates regions where the model's attention is spread out, signaling uncertainty or ambiguity in detection.

This multiplicative combination allows us to prioritize regions that are both \textbf{highly attended} (indicating importance) and \textbf{highly uncertain} (indicating areas where refinement could provide substantial improvement). The regions that exhibit both high attention intensity and high entropy are often small, cluttered, or occluded objects where the model is uncertain but still focused on potential features.

\paragraph{Selection of Top-K Windows}
Once all candidate regions are scored, we select the top-K windows (default \( K = 5 \)) for re-detection. The number \( K \) controls the computational budget, ensuring that only the most informative regions are processed further. The number of selected windows is capped to avoid excessive computation on non-ambiguous regions, thus maintaining the efficiency of the process. In practice, the value of \( K \) ensures that we can focus on regions with the highest potential for improving detection without overwhelming the system with too many redundant candidates.

\paragraph{Efficient Resource Allocation}
The selective routing of computation to high-entropy, high-attention regions ensures that \textbf{ViCrop-Det} operates efficiently, focusing computational resources on the most challenging parts of the image. This targeted refinement allows the model to devote more attention to small or occluded objects that are likely to benefit from re-detection at a higher resolution, without wasting resources on regions that are already well-identified. By prioritizing the \textbf{most uncertain and informative regions}, ViCrop-Det improves the model's accuracy in small-object detection without incurring prohibitive computational costs.

In addition, by focusing only on a small set of candidate windows, we maintain the model's real-time performance. The computational overhead is further minimized because the global gating mechanism described in the main paper often skips images that do not need refinement. As a result, only images with significant uncertainty or ambiguity undergo the additional re-detection process, reducing the number of times the top-K windows are selected.

\subsection{Score Fusion and Final Prediction}

After performing high-resolution re-detection on the selected windows, we combine the refined predictions with the original model outputs to produce the final detection results. This step is crucial for ensuring that the model benefits from the high-resolution reprocessing without compromising the accuracy of the initial detection. The process of fusing these predictions is achieved through two main components: \textbf{class-wise non-maximum suppression (NMS)} and a \textbf{weighted score combination}.

\paragraph{Class-wise Non-Maximum Suppression (NMS)}
To begin the fusion process, we first apply \textbf{class-wise NMS} to remove any redundant detections. NMS is a standard technique in object detection that eliminates overlapping bounding boxes based on a threshold for the intersection-over-union (IoU) metric. However, in \textbf{ViCrop-Det}, we apply NMS \textbf{separately for each class}, ensuring that we do not suppress valuable detections of different object categories that might overlap in spatial locations.

For each object class, NMS operates on the candidate bounding boxes and removes those that overlap with others above a certain IoU threshold (usually \(0.5\)). This step ensures that only the most confident predictions are retained, helping to reduce false positives while preserving the integrity of the original detections, especially for smaller or ambiguous objects.

\paragraph{Weighted Score Fusion}
Once we have the refined predictions from the high-resolution re-detection and the original model's predictions, we combine them to form the final prediction. The combination is performed using a \textbf{weighted score fusion} method, where the final score for each detection is computed as a weighted average of the initial model score \( s_o \) and the refined detection score \( s_r \):

\begin{equation}
s_f = \alpha s_o + (1 - \alpha) s_r
\end{equation}

Here, \( s_f \) is the final fused score, \( s_o \) is the original detection score, \( s_r \) is the score of the refined detection, and \( \alpha \) is a weight parameter that controls the contribution of each score to the final prediction. In our experiments, we empirically set \( \alpha = 0.7 \), which allows the final prediction to primarily rely on the original detection's confidence while also incorporating the refined detection's score to correct missed or under-confident small objects. 

The choice of \( \alpha = 0.7 \) ensures that the fusion process does not overly rely on the re-detected crops, which are intended to correct errors in small-object detection, but instead maintains the precision of the original model detections. This careful balance prevents introducing too much noise from the refined detections while still leveraging the information that comes from focusing on high-entropy regions.

\paragraph{Rationale Behind the Weighted Fusion}
The fusion of original and refined predictions is critical in \textbf{ViCrop-Det}, as it combines the best of both worlds: the initial model's ability to make confident predictions for well-detected objects and the high-resolution refinement to capture fine details of small or ambiguous objects. This strategy ensures that the detection results are both \textbf{precise} and \textbf{robust}, particularly for challenging small-object detection tasks.

By using a weighted score fusion mechanism, \textbf{ViCrop-Det} ensures that \textbf{small objects}, which often lack enough discriminative features for reliable detection, are not completely disregarded. Instead, the model corrects its confidence for these small or occluded objects based on the refined detections, which are targeted at high-entropy regions where the model was uncertain. This process significantly \textbf{improves recall} for small objects without significantly sacrificing the \textbf{precision} of larger objects.

\subsection{Computational Efficiency}

ViCrop-Det introduces minimal additional computation overhead, primarily due to the targeted reprocessing of high-entropy regions. Unlike traditional methods, which might require reprocessing the entire image or large portions of it, \textbf{ViCrop-Det} focuses only on regions where the model is uncertain---i.e., regions with high spatial attention entropy (SAE). This selective re-detection ensures that computational resources are utilized only where necessary, preventing unnecessary computation on well-detected or confident regions.

The \textbf{overall complexity} of the approach is effectively controlled by two main factors:
\begin{itemize}
    \item \textbf{Global Gate Mechanism:} The global gate in \textbf{ViCrop-Det} serves as a filter to determine whether an image requires additional refinement. If the model's initial detection pass is already confident, the gate prevents any further processing. The decision to proceed with further re-detection is based on a combination of the \textbf{global entropy threshold} ($\tau_g$) and the \textbf{mean attention threshold} ($\mu$) across the entire image. If the entropy value of the image is low and the model's attention is well-focused (i.e., high certainty), the image is skipped entirely, saving computational resources. This \textbf{early-exit} mechanism ensures that the approach incurs overhead only on images that truly benefit from reprocessing, improving both speed and efficiency.
    \item \textbf{Cap on the Number of Selected Windows:} Another crucial factor controlling the complexity is the \textbf{limit on the number of selected windows} ($K$) for re-detection. As discussed in the \textbf{Targeted Region Selection and Cropping} section, only the top-$K$ high-entropy windows are selected for further reprocessing. The value of $K$ is typically set to $K = 5$, ensuring that no more than 5 candidate regions are re-detected at a higher resolution. By limiting the number of windows, \textbf{ViCrop-Det} avoids excessive computation on small, less informative regions. The selection of a small, fixed number of regions strikes a balance between computational efficiency and detection improvement, ensuring that only the most ambiguous parts of the image are reprocessed.
\end{itemize}

\paragraph{Efficiency vs. Performance Trade-off}
Despite the added complexity of the high-entropy region selection and high-resolution re-detection, \textbf{ViCrop-Det} introduces only a modest increase in runtime. This is due to the targeted nature of the refinement process, where computation is focused on the most uncertain areas. Since the majority of images (particularly those with well-focused regions) do not require re-detection, the global gate often skips these images entirely, further minimizing computational overhead.

The use of \textbf{multi-scale cropping} to refine high-entropy regions (with scales of $0.25$, $0.5$, and $0.75$) allows \textbf{ViCrop-Det} to efficiently capture small objects at different scales without the need for exhaustive reprocessing at every scale. Only the top-$K$ windows (default $K=5$) are selected for re-detection, ensuring that the approach maintains both \textbf{high accuracy for small objects} and \textbf{efficiency} by limiting the number of additional high-resolution forward passes.

\paragraph{Real-time Applicability}
The \textbf{computational efficiency} of \textbf{ViCrop-Det} makes it suitable for real-time applications, such as autonomous driving, surveillance, or remote sensing, where both accuracy and speed are critical. The ability to selectively reprocess regions of the image without re-running the entire model or overloading the system with redundant computations ensures that \textbf{ViCrop-Det} can be deployed in time-sensitive environments while maintaining high detection performance.

In practical terms, the framework demonstrates only a \textbf{20--23\% overhead} in throughput compared to the baseline models, as shown in the experimental results. This overhead is primarily due to the multi-scale cropping and re-detection of high-entropy regions, but the performance gains in small-object detection far outweigh the modest increase in runtime.

\end{document}